\documentclass{article} %
\usepackage[preprint]{colm2025_conference}

\usepackage{microtype}
\usepackage{hyperref}
\usepackage{url}
\usepackage{booktabs}

\usepackage{lineno}

\definecolor{darkblue}{rgb}{0, 0, 0.5}
\hypersetup{colorlinks=true, citecolor=darkblue, linkcolor=darkblue, urlcolor=darkblue}

\usepackage{multirow}
\usepackage{amssymb}
\usepackage{amsmath}
\usepackage{xcolor}
\usepackage{colortbl}
\usepackage{graphicx}
\usepackage{makecell}

\usepackage{pifont}%
\usepackage[most]{tcolorbox}
\usepackage{listings}
\usepackage{fontawesome5}

\newcommand{\cmark}{\ding{51}}%
\newcommand{\xmark}{\ding{55}}%
\usepackage{lineno}
\usepackage{hyperref}

\title{OmniVox: Zero-Shot Emotion Recognition with Omni-LLMs}

\author{
    John Murzaku\textsuperscript{$1,3$},
    Owen Rambow\textsuperscript{{$2,3$}} \\
    \textsuperscript{$1$}Department of Computer Science \\
    \textsuperscript{$2$}Department of Linguistics\\
    \textsuperscript{$3$}Institute for Advanced Computational Science\\
    Stony Brook University\\
    \texttt{jmurzaku@cs.stonybrook.edu}
}

\begin{document}

\ifcolmsubmission
\linenumbers
\fi

\maketitle
\begin{abstract}
The use of omni-LLMs (large language models that accept any modality as input), particularly for multimodal cognitive state tasks involving speech, is understudied. We present OmniVox, the first systematic evaluation of four omni-LLMs on the zero-shot emotion recognition task. We evaluate on two widely used multimodal emotion benchmarks: IEMOCAP and MELD, and find zero-shot omni-LLMs outperform or are competitive with fine-tuned audio models. Alongside our audio-only evaluation, we also evaluate omni-LLMs on text only and text and audio. We present acoustic prompting, an audio-specific prompting strategy for omni-LLMs which focuses on acoustic feature analysis, conversation context analysis, and step-by-step reasoning. We compare our acoustic prompting to minimal prompting and full chain-of-thought prompting techniques. We perform a context window analysis on IEMOCAP and MELD, and find that using context helps, especially on IEMOCAP. We conclude with an error analysis on the generated acoustic reasoning outputs from the omni-LLMs. 
\end{abstract}
\section{Introduction}
The Emotion Recognition in Conversation (ERC) task, a subset of broader cognitive state modelling tasks, has received a significant amount of attention from both the NLP and speech processing communities. Many corpora have been created, involving speech, text transcripts, and videos as modalities, and these corpora have been used to explore unimodal or multimodal architectures. The previous works often report results, usually fine-tuned and tested on one modality (mainly text or speech); however, these studies do not have a unifying approach. In other words, is there a task-agnostic, modality-agnostic, generalized model that can perform well on the ERC task
in a zero-shot manner?  Recently, there has been an emergence of omni-LLMs such as Gemini \citep{team2024gemini, gemini}, GPT-4o \citep{openai}, and Phi-4-Multimodal \citep{phi4}, which are models that accept any modality as input, and either output a single modality or combination of modalities. On March 26th, 2025, Qwen-2.5-Omni, a new omni-LLM was released \citep{qwen25omni7b}, which we do not evaluate on due to its recent release date. This model is the successor to Qwen-Audio \citep{chu2023qwen} and Qwen-2-Audio \citep{chu2024qwen2}, whose multi-task pre-training included emotion recognition data.

Our study aims to discover the emergent audio capabilities of omni-LLMs for ERC. While omni-LLMs inherently accept multiple modalities (text, speech, image) as input, we prioritize audio-to-text inference, which specifically translates raw audio inputs directly into text emotion labels. We emphasize that this is a previously unexplored direction. We also present results on additional modalities, including text and combined text-speech inputs as complementary analyses to further enrich our findings.

Our main contributions are summarized as follows:
\begin{itemize}
    \item To the best of our knowledge, we are the first to perform a systematic evaluation on four omni-LLMs for
    zero-shot emotion recognition from audio only. Compared to previous audio-only baselines, we find that our zero-shot method outperforms state-of-the-art fine-tuned audio model baselines (up to 7.6\% improvement for IEMOCAP, 4.7\% improvement for MELD), or are very competitive compared to baselines (2.9\% lower for IEMOCAP and 0.7\% lower for MELD).
    We also evaluate on text only and both text and speech as inputs.
    
    \item We present results on how to prompt omni-LLMs with audio input for the emotion task with three different zero-shot prompting strategies: (i) minimal, where we instruct the model to only predict an emotion; (ii) Acoustic, where we instruct the model to perform an acoustic analysis then predict the emotion; (iii) chain-of-thought (CoT), where we instruct the model to perform an acoustic analysis, perform a step-by-step reasoning, and then output the label. 
    
    \item We present results on how many turns of audio context helps. We specifically report results on no context, and various contexts, up to 12 turns. Our results serve as a preliminary analysis into how many turns of dialogues omni-LLMs can track. 
    
    \item We perform a comprehensive error analysis on where audio models fail for emotion, particularly focusing on acoustic reasoning. We conclude with insights for future emotion tasks, and present a general discussion on how to best use omni-LLMs. 
\end{itemize}

\section{Related Work}

\textbf{Fine-tuned Models for ERC} Many papers have focused on fine-tuned multimodal (text, audio, vision) models for ERC. Most recently, there has been multiple papers \citep{shou2024revisiting, zhao2025temporal} achieving state-of-the-art (SOTA) results on IEMOCAP and MELD using multimodal
state space models, specifically the Mamba architecture \citep{gu2023mamba}. Similarly, with the exploding popularity of large language models (LLMs), previous papers have fine-tuned LLMs, \citep{lei2023instructerc, wu2024beyond} used LLM supervised pretraining \citep{dutta2025llm}, all yielding SOTA results when the papers released. Finally, the majority of recent ERC work optimizes for fusion network architectures: teacher-student fusion networks \citep{yun-etal-2024-telme}, graph neural network (GNN) based architectures \citep{meng2024revisiting}, feature alignment fusion networks \citep{wang2025enhancing}, multimodal transformer fusion networks \citep{cheng2024emotion,sasu2025akan}, and early/late fusion with TTS generated emotional speech \citep{soubki2025synthetic}.

\textbf{Zero-shot ERC} Conversely, there has not been as much attention for the zero-shot ERC task; for the work that exists, the focus is only on the text modality. Most recently, \citet{wu2024beyond} perform zero-shot experiments on IEMOCAP and MELD with Claude-3.5 Sonnet \citep{claude}. However, this is on the text-only modality, with descriptions of speech features in natural language. \citet{lei2023instructerc} similarly performed zero-shot experiments with GPT-3.5 \citep{gpt35}, but only on text transcripts of IEMOCAP and MELD. 

Our work distinguishes itself from previous research in two salient ways: first, we use a multimodal framework with a unifying omni-LLM, using text, audio, or both text and audio (with particular emphasis placed on audio). Second, and most critically, our entire approach is zero-shot, with a single modality-agnostic omni-LLM. This zero-shot multimodal paradigm positions our approach uniquely within the ERC literature: we are the first to explore the zero-shot ERC task with audio only, and we are the first to offer insights and a framework for working with omni-LLMs.

\begin{figure}[!htbp]
    \centering
    \includegraphics[width=\linewidth]{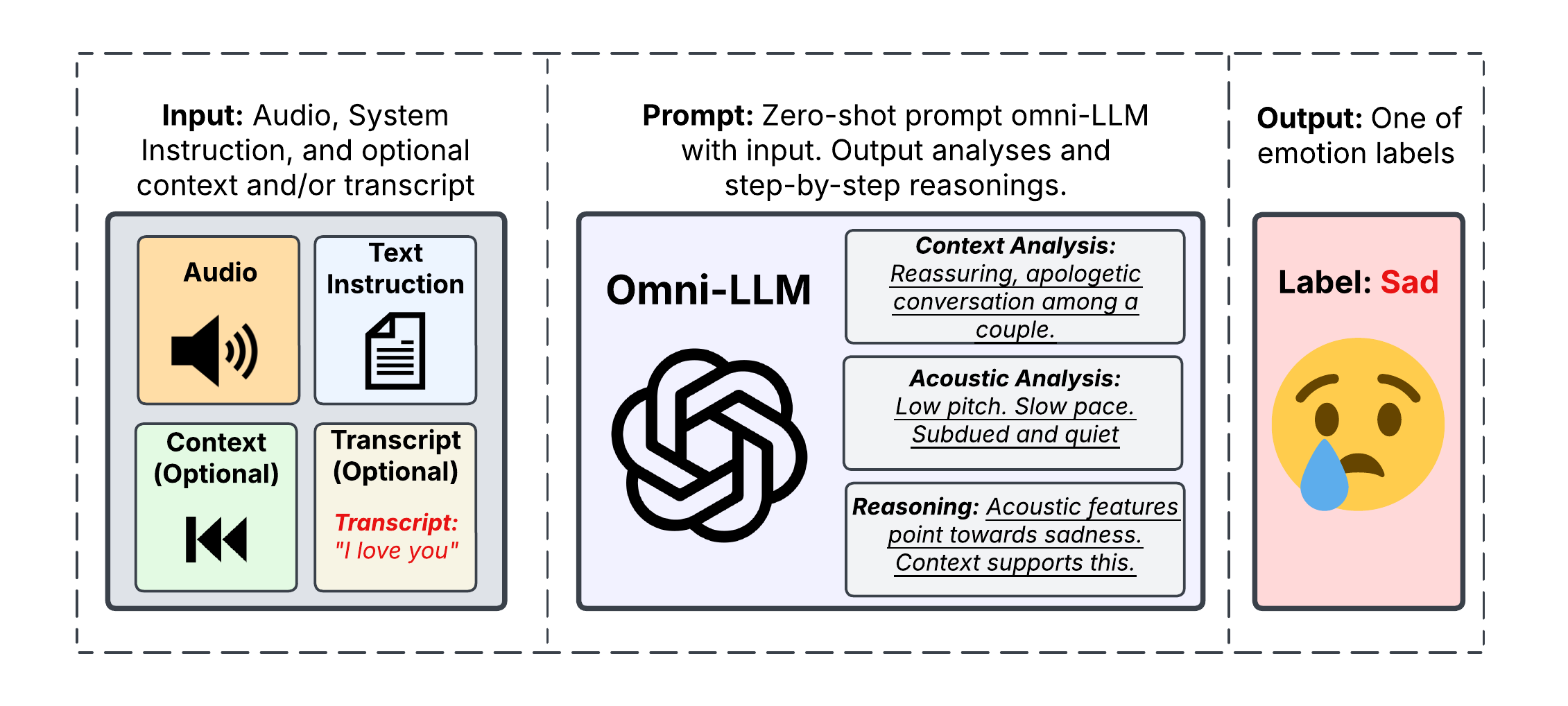}
    \caption{The proposed \textbf{OmniVox} framework. We perform zero-shot emotion recognition from audio inputs enhanced by text instructions, and optional contextual information or transcripts. We then generate a context analysis, acoustic feature interpretation, and a final chain-of-thought reasoning, ultimately predicting a specific emotion label (e.g., sad in this example).}
    \label{fig:system}
\end{figure}

\section{Methodology}
\subsection{Datasets} We use two widely used multimodal emotion recognition corpora:
\textbf{IEMOCAP} We use the IEMOCAP (Interactive Emotional Dyadic Motion Capture) corpus \citep{busso2008iemocap}, which consists of nine emotion labels: neutral, anger, frustration, happiness, excitement, sadness, fear, surprise, and other. The corpus consists of dyadic conversations among paid actors and is annotated on the utterance level. Regarding modalities, the corpus includes video, speech, motion capture of face, and text transcriptions; we only use the speech and the text transcriptions. Regarding labels, we collapse the label set to six labels (anger, happiness, excitement, sadness, frustration, and neutral), following the most recent work using IEMOCAP \citep{gong-etal-2024-mapping,wu2024beyond}.

\textbf{MELD} We also use the MELD (Multimodal EmotionLines Dataset) corpus \citep{poria2018meld}, which contains naturally occurring dialogues from the TV series \textit{Friends}. The corpus is annotated with seven emotion categories: anger, disgust, fear, joy, neutral, sadness, and surprise. Similar to IEMOCAP, MELD also includes speech, video, and text transcriptions as modalities, and we again only use the speech and the text transcriptions. 

\subsection{Models}
We provide a brief overview of the omni-LLM models that we use in this paper. A summary of the supported input and output modalities of each model is shown in Appendix Table~\ref{tab:model-modalities}.

\textbf{Gemini 2.0} We use the Gemini-2.0-Flash (henceforth Gemini) and the Gemini-2.0-Flash-Lite (henceforth Gemini-Lite) models through the Google Gemini API \citep{gemini}. The Gemini-2.0 series of models are optimized for multimodal understanding across many tasks spanning text, audio, video, and image, outperforming the Gemini-1.5 models \cite{team2024gemini}. We also perform experiments on Gemini-Lite, which provides faster runtimes and cost efficiency. Table~\ref{tab:model-modalities} shows the modalities of both models: both Gemini and Gemini-Lite accept any modality as input, but Gemini outputs any modality (with future API releases allowing audio), while Gemini-Lite only allows text output. 

\textbf{GPT-4o (audio)} We perform experiments on the audio checkpoint of GPT-4o \citep{openai}, specifically \texttt{Gpt-4o-Audio-preview}. While GPT-4o is reported as an omni model, the audio modality has a distinct API endpoint (that is, we cannot use the standard GPT-4o endpoint). Table~\ref{tab:model-modalities} shows the allowed input modalities: GPT-4o accepts audio and/or text (with mandatory audio input), and can output either text, audio, or text and audio. 

\textbf{Phi-4-Multimodal} We use the newly released \texttt{Phi-4-Multimodal-Instruct} model \citep{phi4}. The model consists of 5.6B parameters, with Phi-4-Mini-Instruct as the language model, and separate vision and speech encoders. Table~\ref{tab:model-modalities} shows the allowed input modalities: Phi-4-Multimodal-Instruct accepts audio, text, or image (or any combination of the three), and only outputs text.

\subsection{Task Setup} 
We describe our task setup shown in Figure~\ref{fig:system}. 

\textbf{Step 1: Input.} We begin by building our inputs to the omni-LLM. We begin with our audio file as input, followed by the text instruction (we describe the text instruction in the following subsection). Optionally, we include audio and/or text context (up to 12 previous turns of dialogue), and an optional text transcript. 

\textbf{Step 2: Prompt.} We prompt the omni-LLM with the input from Step 1. For our main prompt, the omni-LLM then generates a context analysis of the conversation (if context is provided), an acoustic analysis detailing features such as pitch, pace, and volume, and a step-by-step reasoning. We discuss further prompts and strategies in Section~\ref{sec:prompts}.

\textbf{Step 3: Output.} The final output consists of one of the corpus specific emotion labels. We note that this final label comes after the analyses (and is therefore outputted at the end of the one single LLM call in Step 2).

In all of our main results (Section~\ref{sec:main-results}), we use a chain-of-thought (CoT) prompting approach, with the OmniVox framework mentioned above. Further, for most of our experiments, we follow a conversation context window of three turns (c=3), following \citet{wu2024beyond}, who used the last three turns of context with text acoustic features. 

\subsection{Prompting Strategies}
\label{sec:prompts}
We provide a brief overview of the three prompting strategies we use. We include the full prompts in Appendix~\ref{appendix:prompts}. We remind the reader that our prompts are system instructions given to the omni-LLMs, and the omni-LLMs primarily take in an audio utterance as input (and optionally: audio context, text context, or a text transcript of the utterance).

\textbf{Minimal} Our minimal prompt aims to classify emotion in the simplest way possible. To that end, our prompt consists of a simple instruction asking to classify the emotion in the audio and a list of which labels to use.

\textbf{Acoustic} Our Acoustic prompt adds an initial reasoning step to our model. Specifically, we ask the model to first provide an acoustic analysis of the audio, then predict the final emotion label.

\textbf{Chain-of-Thought (CoT)} Our CoT prompt uses chain-of-thought prompting \citep{wei2022chain}, explicitly instructing the model to reason step-by-step. We first include the instruction to perform an
acoustic analysis from our Acoustic prompt, and then add an instruction to perform a final CoT reasoning step.

\subsection{Evaluation} We perform a weighted-F1 (W-F1) evaluation for both the test sets of IEMOCAP and MELD following the most recent works \citep{wu2024beyond, yun-etal-2024-telme, shou2024revisiting}. We note that MELD contains an imbalanced label distribution, with a majority label of neutral (about 48.1\% of the test set). We include an expanded confusion analysis in Section~\ref{sec:error-analysis}. 

\section{Experiments}
\label{sec:experiments}
We split our experiments into four subsections: Section~\ref{sec:main-results} begins with results on each of the modalities (audio, text, text+audio). Section~\ref{sec:prompting} focuses on prompting strategies for omni-LLMs, exclusively using the audio modality. Section~\ref{sec:context} performs an analysis on how many turns of context help for audio-only experiments. Finally, Section~\ref{sec:comparison_to_prev} compares OmniVox's results on all modalities to previously reported results.
\begin{table*}[!htbp]
\small
\centering
\begin{tabular}{l l | c c | c c | c c || c}
\toprule
\textbf{Dataset} & \textbf{Model} & \multicolumn{2}{c|}{\textbf{Audio}} & \multicolumn{2}{c|}{\textbf{Text}} & \multicolumn{2}{c||}{\textbf{Text + Audio}} & \multicolumn{1}{c}{\textbf{Gold Feats.}} \\
& & \textbf{c=0} & \textbf{c=3} & \textbf{c=0} & \textbf{c=3} & \textbf{c=0} & \textbf{c=3} & \\
\midrule
\multirow{4}{*}{IEMOCAP} 
  & Gemini           & 45.6 & \cellcolor[HTML]{FFF8E6}\textbf{49.9} & 39.2 & 44.2 & 45.9 & 46.3 & 52.7 \\
  & Gemini Lite      & 41.2 & \cellcolor[HTML]{FFF8E6}\textbf{51.5} & 37.0 & 42.5 & 42.1 & 46.1 & 49.9 \\
  & Gpt-4o-Audio     & 45.8 & \cellcolor[HTML]{FFF8E6}\textbf{51.8} & --   & --   & 44.7 & 47.6 & -- \\
  & Phi-4-Multimodal & 35.0 & 37.6 & --   & --   & 37.9 & \cellcolor[HTML]{FFF8E6}\textbf{43.1} & -- \\
\midrule
\multirow{4}{*}{MELD} 
  & Gemini           & 43.7 & 48.9 & 61.2 & \cellcolor[HTML]{FFF8E6}\textbf{62.8} & 58.5 & 61.7 & 54.1 \\
  & Gemini Lite      & 44.2 & 47.8 & 59.7 & 60.4 & 57.6 & \cellcolor[HTML]{FFF8E6}\textbf{61.8} & 42.5 \\
  & Gpt-4o-Audio     & 51.3 & 47.7 & --   & --   & 61.0 & \cellcolor[HTML]{FFF8E6}\textbf{61.7} & -- \\
  & Phi-4-Multimodal & 36.3 & 35.3 & --   & --   & \cellcolor[HTML]{FFF8E6}\textbf{43.4} & 43.1 & -- \\
\bottomrule
\end{tabular}
\caption{Weighted-F1 (W-F1) performance of four omni-LLMs on the IEMOCAP and MELD datasets. Results are shown for four modalities: \textbf{Audio} (c=0, c=3), \textbf{Text} (c=0, c=3), combined \textbf{Text + Audio} (c=0, c=3), and \textbf{Gold Feats.} (gold acoustic textual descriptions from \citet{wu2024beyond}). Highlighted cells highlight the best performing configuration for each model.}
\label{tab:main-results}
\end{table*}

\subsection{Multimodal Results}
\label{sec:main-results}
We describe the multimodal results per corpus. We use the chain-of-thought prompt mentioned in Section~\ref{sec:prompts}. Our results are shown in Table~\ref{tab:main-results}. We also show examples of our LLM outputs
from IEMOCAP and MELD in Appendix~\ref{appendix:examples}.
\paragraph{IEMOCAP.}   Audio-only, with context (c=3) results beat all other modalities for IEMOCAP, for three models.  Phi-4 is an outlier, consistently performing worse than the other models under all conditions, and showing a different pattern for its best result. 

\textbf{Audio}  GPT-4o-audio with context (c=3) achieves the highest W-F1 (51.8) over all conditions and models,
outperforming Gemini Lite (51.5) and Gemini (49.9). In all cases for the audio modality besides Phi-4, we find context helps. However, with no-context conditions (c=0), performance for GPT-4o-audio and Gemini is comparable (45.8 and 45.6, respectively).

\textbf{Text}
Both Gemini and Gemini Lite show large gains with context added compared to no context (increase of 5.0\% for Gemini, and 5.5\% for Gemini-Lite) . However, our text-only results still are below the audio results. 

\textbf{Text+Audio} GPT-4o-audio performs best with context (c=3), achieving a W-F1 score of 47.6, clearly benefiting from context integration. However, it performs worse than the audio-only modality than some models with no context (c=0), and all models with context (c=3).

\paragraph{MELD.}  For MELD, unlike IEMOCAP, the text modality did increase the performance, with Gemini achieving the highest overall W-F1 score on \textbf{Text} only, and Gemini-Lite and Gpt-4o-Audio performing best on \textbf{Text+Audio}, all with context (c=3).  Again, Phi-4 is an outlier, consistently performing worse than the other models, and again showing a different pattern for its best result.

\textbf{Audio} Unlike IEMOCAP, GPT-4o-audio achieves the highest W-F1 (51.3) in the no-context condition (c=0), substantially outperforming both Gemini (43.7) and Gemini Lite (44.2). Interestingly, when context is added (c=3), GPT-4o-audio's performance decreases (to 47.7) while Gemini improves (to 48.9).

\textbf{Text} 
Gemini achieves the highest overall W-F1 score (62.8) with context (c=3), showing a modest improvement (1.6\%) over the no-context condition. Gemini consistently outperforms Gemini Lite in both conditions. 

\textbf{Text+Audio} 
Gemini-Lite performs better than text-only with context (1.4\% increase) and audio-only with context (14\% increase) and Gpt-4o-Audio performs better than audio-only by a large margin (14\% increase). 

\subsection{Prompting Strategies}
\label{sec:prompting}
\definecolor{checkmarkcolor}{HTML}{2E8B57} %
\definecolor{xmarkcolor}{HTML}{B22222} %
\definecolor{lightgray}{HTML}{FFF8E6} %
\definecolor{darkgray}{HTML}{FFF2CC} %
\definecolor{highlightcell}{HTML}{FFF2CC} %

\newcommand{\better}[1]{\cellcolor{lightgray}\textbf{#1}}
\newcommand{\improveAcoustic}[1]{\cellcolor{lightgray}\textbf{#1}}
\newcommand{\improveCoT}[1]{\cellcolor{darkgray}\textbf{#1}}
\newcommand{\contextNo}[1]{\cellcolor{xmarkcolor!25}{#1}}
\newcommand{\contextYes}[1]{\cellcolor{checkmarkcolor!25}{#1}}

\begin{table*}[!htbp]
\setlength\tabcolsep{2pt} %
\small
\centering
\begin{tabular}{lc|ccc|ccc}
\toprule
\multirow{2}{*}{\textbf{Model}} & 
\multirow{2}{*}{\textbf{Context}} & 
\multicolumn{3}{c|}{\textbf{IEMOCAP}} & 
\multicolumn{3}{c}{\textbf{MELD}} \\
\cmidrule(lr){3-5}\cmidrule(lr){6-8}
& & \textbf{Minimal} & \textbf{Acoustic} & \textbf{CoT} & \textbf{Minimal} & \textbf{Acoustic} & \textbf{CoT} \\
\midrule
\multirow{2}{*}{Gemini} & \contextNo{\xmark} & 45.0 & \improveAcoustic{48.1} & 41.2 & 35.8 & \improveAcoustic{42.5} & \improveCoT{43.7} \\
                       & \contextYes{\cmark} & 51.7 & \improveAcoustic{51.9} & 49.9 & {48.6} & 46.2 & \improveCoT{48.9} \\
\midrule
\multirow{2}{*}{Gemini-Lite} & \contextNo{\xmark} & 42.7 & \improveAcoustic{43.5} & 41.2 & 41.0 & \improveAcoustic{48.2} & 44.2 \\
                          & \contextYes{\cmark} & 48.8 & 44.6 & \improveCoT{51.5} & 40.5 & \improveAcoustic{45.9} & \improveCoT{47.8} \\
\midrule
\multirow{2}{*}{GPT-4o-Audio} & \contextNo{\xmark} & 47.0 & \improveAcoustic{48.4} & 45.8 & 31.7 & \improveAcoustic{50.5} & \improveCoT{51.3} \\
                           & \contextYes{\cmark} & 47.5 & \improveAcoustic{50.5} & \improveCoT{51.8} & 31.2 & \improveAcoustic{45.7} & \improveCoT{47.7} \\
\midrule
\multirow{2}{*}{Phi-4-Multimodal} & \contextNo{\xmark} & 15.8 & \improveAcoustic{29.2} & \improveCoT{35.0} & 21.8 & \improveAcoustic{31.5} & \improveCoT{36.3} \\
                               & \contextYes{\cmark} & 24.4 & \improveAcoustic{42.6} & \improveCoT{47.6} & 30.6 & \improveAcoustic{34.7} & 35.3 \\
\bottomrule
\end{tabular}
\caption{Weighted F1 scores (\%) for different models and prompting structures (Minimal, Acoustic, Chain-of-Thought) on IEMOCAP and MELD datasets. "\xmark" indicates no conversation context (c=0), and "\cmark" indicates including conversation context (c=3). Highlighted cells indicate performance improvement from Minimal to Acoustic (light yellow), and from Acoustic to CoT (dark yellow).}
\label{tab:prompting_structures}
\end{table*}

Table~\ref{tab:main-results} emphasizes that the \textbf{Gold Feats.} column, which uses the text descriptions of gold acoustic features from \citet{wu2024beyond}, achieves better or competitive results compared to audio-only. We aim to investigate the following question: does explicitly prompting for acoustic features help omni-LLMs for the audio-only modality?
We present results for three different prompting strategies Minimal, Acoustic, and Chain-of-Thought (CoT) in Table~\ref{tab:prompting_structures}.

\textbf{IEMOCAP} We find that explicitly prompting for acoustic analyses consistently enhanced model performance compared to minimal prompts. Specifically, Gemini achieved notable improvements from 45.0\% to 48.1\% (without context) and from 51.7\% to 51.9\% (with context). Similarly, GPT-4o-Audio improved its weighted F1 scores from 47.0\% to 48.4\% (without context) and 47.5\% to 50.5\% (with context). Phi-4-Multimodal yielded the largest performance gain gains, particularly when context was included, improving from 24.4\% (minimal) to 42.6\% (acoustic). Our CoT prompting method further enhanced Gemini-Lite and GPT-4o-Audio performances when context was provided, reaching 51.5\% and 51.8\%  respectively.

\textbf{MELD} Similar to our results for IEMOCAP, prompting for acoustic analyses also led to consistent improvements across models, especially without context. Gemini's weighted F1 score increased markedly from 35.8\% to 42.5\%, and GPT-4o-Audio yielded substantial gains from 31.7\% to 50.5\% in acoustic conditions. Our CoT prompting further improved performance, with GPT-4o-Audio achieving 51.3\% without context and Gemini attaining 48.9\% when context was included. Phi-4 displayed consistent improvement across prompt types, moving from 21.8\% (minimal) to 36.3\% (CoT) without context.

\textbf{Summary} Our results suggest that acoustic information contributes to emotion recognition accuracy across both IEMOCAP and MELD. Further, CoT generally enhances performance, especially when contextual information is provided.

\subsection{Context Window}
\begin{table*}[!htbp]
\setlength\tabcolsep{3pt}
\centering
\small
\begin{tabular}{l l | ccccccc}
\toprule
\textbf{Dataset} & \textbf{Model} & \multicolumn{7}{c}{\textbf{Context Window Size (c)}} \\
\cmidrule(lr){3-9}
& & \textbf{0} & \textbf{1} & \textbf{2} & \textbf{3} & \textbf{4} & \textbf{5} & \textbf{12} \\
\midrule
\multirow{4}{*}{IEMOCAP}
& Gemini & \textcolor{blue}{45.9} & \cellcolor[HTML]{D9EAD3}47.7 & \cellcolor[HTML]{D9EAD3}48.5 & \cellcolor[HTML]{D9EAD3}49.8 & \cellcolor[HTML]{D9EAD3}53.1 & \cellcolor[HTML]{D9EAD3}50.6 & \cellcolor[HTML]{D9EAD3}52.7 \\
& Gemini-Lite & \textcolor{blue}{41.3} & \cellcolor[HTML]{D9EAD3}45.7 & \cellcolor[HTML]{D9EAD3}47.4 & \cellcolor[HTML]{D9EAD3}51.5 & \cellcolor[HTML]{D9EAD3}51.2 & \cellcolor[HTML]{D9EAD3}53.8 & \cellcolor[HTML]{D9EAD3}54.1 \\
& Gpt-4o-Audio & \textcolor{blue}{\textbf{46.0}} & \cellcolor[HTML]{D9EAD3}\textbf{47.9} & \cellcolor[HTML]{D9EAD3}\textbf{51.1} & \cellcolor[HTML]{D9EAD3}\textbf{51.8} & \cellcolor[HTML]{D9EAD3}\textbf{53.2} & \cellcolor[HTML]{D9EAD3}\textbf{54.2} & \cellcolor[HTML]{D9EAD3}\textbf{55.9} \\
& phi-4-multimodal & \textcolor{blue}{34.8} & 34.3 & \cellcolor[HTML]{D9EAD3}35.8 & \cellcolor[HTML]{D9EAD3}37.6 & \cellcolor[HTML]{D9EAD3}37.7 & \cellcolor[HTML]{D9EAD3}40.5 & \cellcolor[HTML]{D9EAD3}40.9 \\
\midrule
\multirow{4}{*}{MELD}
& Gemini & \textcolor{blue}{43.2} & \cellcolor[HTML]{D9EAD3}46.7 & \cellcolor[HTML]{D9EAD3}\textbf{48.0} & \cellcolor[HTML]{D9EAD3}\textbf{48.8} & \cellcolor[HTML]{D9EAD3}47.7 & \cellcolor[HTML]{D9EAD3}47.9 & \cellcolor[HTML]{D9EAD3}47.3 \\
& Gemini-Lite & \textcolor{blue}{43.6} & \cellcolor[HTML]{D9EAD3}\textbf{48.6} & \cellcolor[HTML]{D9EAD3}47.9 & \cellcolor[HTML]{D9EAD3}47.7 & \cellcolor[HTML]{D9EAD3}47.5 & \cellcolor[HTML]{D9EAD3}\textbf{48.3} & \cellcolor[HTML]{D9EAD3}\textbf{48.0} \\
& Gpt-4o-Audio & \textcolor{blue}{\textbf{51.3}} & 48.3 & 48.2 & 47.4 & \textbf{47.8} & 48.0 & 47.1 \\
& phi-4-multimodal & \textcolor{blue}{36.3} & \cellcolor[HTML]{D9EAD3}37.2 & 35.9 & 35.3 & \cellcolor[HTML]{D9EAD3}36.9 & 36.4 & \cellcolor[HTML]{D9EAD3}37.0 \\
\bottomrule
\end{tabular}
\caption{Effect of context window size on audio-only performance (W-F1 scores). Baseline (c=0) values are shown in blue, green highlights indicate performance improvements over the no context baseline, and bold indicates the best performance in each column.}
\label{tab:context_window_analysis}
\end{table*}

\label{sec:context}
We aim to answer the following question: how many turns of \textit{audio} context help?
In other words, is there an optimal amount of turns for omni-LLMs to understand the emotion and contextual dynamics across a conversation, purely from audio? Table~\ref{tab:context_window_analysis}
shows our per-corpus results, using the CoT style prompt.
We use context windows from ${0-5, 12}$: our maximum window choice mirror \citet{wu2024beyond}, who use context windows of 12 turns.  

\textbf{IEMOCAP} Our results show that incorporating audio context enhances the emotional recognition performance for IEMOCAP across all evaluated models. Gemini improves from our c=0 baseline of 45.9\% to 53.1\% with a context window of 4 utterances. Gemini-Lite and Gpt-4o-Audio show similar trends, achieving peak performances of 54.1\% and 55.9\% at context windows of 12, respectively.

\textbf{MELD} The effectiveness of audio context varies between models. While Gemini and Gemini-Lite exhibit clear performance improvements when context is introduced—reaching peak performance at 48.8\% and 48.6\% respectively, Gpt-4o-Audio shows a decline from its baseline of 51.3\% when context is introduced. Phi-4-multimodal shows only marginal and inconsistent improvements across different context sizes. We hypothesize that this could be due to omni-LLMs failing with more complex audio: MELD has multi-party TV show dialogues with background noise, audience laughter, and varying conditions, whereas IEMOCAP has better quality, dyadic audio recordings in a lab setting. We quantify this using the signal-to-noise ratio (SNR) and find that MELD and IEMOCAP have similar SNRs (14.35dB for MELD vs 15.86dB for IEMOCAP), however, MELD shows 56\% higher standard deviation (8.82dB vs 5.64dB).%
\begin{table*}[!htbp]
\centering
\small
\begin{tabular}{lcccc}
\toprule
\textbf{Dataset} & \textbf{Method} & \textbf{Audio} & \textbf{Text} & \textbf{Text + Audio} \\
\midrule
\multirow{5}{*}{IEMOCAP} 
    & \citet{lei2023instructerc}   & --   & 71.4   & -- \\
    & \citet{shou2024revisiting}    & \textbf{58.8} & 65.7   & \textbf{70.2} \\
    & \citet{wu2024beyond}          & \textcolor{red}{52.3} & \textbf{72.6}   & -- \\
    & \citet{yun-etal-2024-telme}          & \textcolor{red}{48.1} & 66.6   & \textbf{69.3} \\
    \rowcolor[HTML]{FFF2CC}
    & \textbf{OmniVox (Ours)}       & 55.9 & 44.2   & 47.6 \\
\midrule
\multirow{5}{*}{MELD} 
    & \citet{lei2023instructerc}   & --   & \textbf{69.2}   & -- \\
    & \citet{shou2024revisiting}    & \textbf{52.0} & {63.9}   & \textbf{65.6} \\
    & \citet{wu2024beyond}          & \textcolor{red}{47.9} & 67.6   & -- \\
    & \citet{yun-etal-2024-telme}          & \textcolor{red}{46.6} & 66.6   & \textbf{67.2} \\
    \rowcolor[HTML]{FFF2CC}
    & \textbf{OmniVox (Ours)}               & 51.3 & 62.8   & 61.7 \\
\bottomrule
\end{tabular}%

\caption{Comparison of our top-performing zero-shot OmniVox results with prior work that fine-tuned text, audio, or multimodal text and audio models~\citep{lei2023instructerc, wu2024beyond, shou2024revisiting, yun-etal-2024-telme}. All values shown are W-F1 scores. Best results in each category are \textbf{bolded}. Fine-tuned results worse than our zero-shot OmniVox are in {\color{red} red}.}
\label{tab:comparison_wu}
\end{table*}

\subsection{Comparison to Previous Work}
\label{sec:comparison_to_prev}
We conclude by comparing our top performing approaches (GPT-4o-audio, c=12 for IEMOCAP, and GPT-4o-audio, c=0 for MELD) to four recent works, which evaluated on audio, text (or, speech features as text), and text and audio. Our results are shown in Table~\ref{tab:comparison_wu}.

For both IEMOCAP and MELD, our approach achieves competitive audio-only results (55.9\% for IEMOCAP, 51.3\% for MELD), surpassing several fine-tuned models (shown in red), although falling behind the highest reported audio result from \citet{shou2024revisiting}.

\definecolor{goldcolor}{HTML}{2E8B57} %
\definecolor{predcolor}{HTML}{B22222} %
\definecolor{highlightstrong}{HTML}{FFDC99} %
\definecolor{highlightmed}{HTML}{FFF2CC} %
\definecolor{lightgray}{gray}{0.92} %
\definecolor{goldcolor}{HTML}{2E8B57} %
\definecolor{predcolor}{HTML}{B22222} %
\definecolor{highlightstrong}{HTML}{FFDC99} %
\definecolor{highlightmed}{HTML}{FFF2CC} %
\definecolor{lightgray}{gray}{0.92} %
\begin{table}[!htbp]
\centering
\small
\setlength{\tabcolsep}{2pt}
\begin{tabular}{lccc| lccc}
\toprule
\multicolumn{4}{c}{\textbf{IEMOCAP}} & \multicolumn{4}{c}{\textbf{MELD}} \\
\cmidrule(lr){1-4} \cmidrule(lr){5-8}
\textbf{\textcolor{goldcolor}{Gold} $\rightarrow$ \textcolor{predcolor}{Pred.}} & \textbf{Vol.} & \textbf{Pitch} & \textbf{Rate} & 
\textbf{\textcolor{goldcolor}{Gold} $\rightarrow$ \textcolor{predcolor}{Pred.}} & \textbf{Vol.} & \textbf{Pitch} & \textbf{Rate} \\
\midrule
\multicolumn{4}{l}{\textcolor{goldcolor}{\textbf{Ang.}} $\rightarrow$ \textcolor{predcolor}{\textbf{Fru.}}} & 
\multicolumn{4}{l}{\textcolor{goldcolor}{\textbf{Joy}} $\rightarrow$ \textcolor{predcolor}{\textbf{Neu.}}} \\
\midrule
Higher & \cellcolor{highlightmed}\textbf{50.0\%} & \cellcolor{highlightmed}\textbf{51.9\%} & \cellcolor{highlightmed}\textbf{51.0\%} &
Higher & 12.3\% & 22.1\% & 0.0\% \\
Lower  & 9.6\%  & 2.9\%  & 5.8\%  &
Lower  & 30.3\% & 16.0\% & 7.7\% \\
Same   & \cellcolor{highlightmed}40.4\% & \cellcolor{highlightmed}44.2\% & \cellcolor{highlightmed}43.3\% &
Same   & \cellcolor{highlightmed}57.4\% & \cellcolor{highlightstrong}62.0\% & \cellcolor{highlightstrong}\textbf{92.3\%} \\
\midrule
\multicolumn{4}{l}{\textcolor{goldcolor}{\textbf{Fru.}} $\rightarrow$ \textcolor{predcolor}{\textbf{Neu.}}} &
\multicolumn{4}{l}{\textcolor{goldcolor}{\textbf{Sad}} $\rightarrow$ \textcolor{predcolor}{\textbf{Neu.}}} \\
\midrule
Higher & 4.5\% & 0.0\% & 2.3\% &
Higher & 19.8\% & 20.5\% & 1.1\% \\
Lower  & \cellcolor{highlightmed}42.0\% & 33.0\% & \cellcolor{highlightmed}\textbf{53.4\%} &
Lower  & 36.4\% & 16.7\% & 16.8\% \\
Same   & \cellcolor{highlightmed}\textbf{53.4\%} & \cellcolor{highlightstrong}\textbf{67.0\%} & \cellcolor{highlightmed}44.3\% &
Same   & \cellcolor{highlightmed}43.8\% & \cellcolor{highlightstrong}62.9\% & \cellcolor{highlightstrong}\textbf{82.1\%} \\
\midrule
\multicolumn{4}{l}{\textcolor{goldcolor}{\textbf{Hap.}} $\rightarrow$ \textcolor{predcolor}{\textbf{Exc.}}} &
\multicolumn{4}{l}{\textcolor{goldcolor}{\textbf{Ang.}} $\rightarrow$ \textcolor{predcolor}{\textbf{Neu.}}} \\
\midrule
Higher & \cellcolor{highlightmed}42.3\% & \cellcolor{highlightstrong}\textbf{69.2\%} & \cellcolor{highlightstrong}\textbf{71.2\%} &
Higher & 19.4\% & 12.9\% & 3.1\% \\
Lower  & 1.9\%  & 0.0\%  & 0.0\%  &
Lower  & 29.5\% & 24.2\% & 6.2\% \\
Same   & \cellcolor{highlightmed}\textbf{53.8\%} & 30.8\% & 28.8\% &
Same   & \cellcolor{highlightmed}51.2\% & \cellcolor{highlightstrong}62.9\% & \cellcolor{highlightstrong}\textbf{90.6\%} \\
\bottomrule
\end{tabular}
\caption{Divergence between omni-LLM predicted acoustic features and reference acoustic features for IEMOCAP and MELD. Percentages indicate how frequently the predicted acoustic features were higher, lower, or unchanged compared to the gold/reference acoustic features. A color gradient highlights trend strength: darker yellow for strong trends ($\geq$60\%), medium yellow for moderate trends (40-60\%), no highlighting for weak trends ($<$40\%).}
\label{tab:divergence}
\end{table}

\section{Analysis}
\label{sec:error-analysis}
We perform an error analysis for the audio only modality on each corpus using the top performing model configuration (GPT-4o-audio with c=12 for IEMOCAP, GPT-4o-Audio, c=0 for MELD). We split our error analysis into three motivating research questions: 

\textbf{RQ1:} Are there specific emotion pairs that omni-LLMs frequently confuse? If yes, are they aligned with confusion patterns from previous work? 

\textbf{RQ2:} In Section~\ref{sec:experiments}, we saw that explicitly prompting for acoustic descriptions help. To what extent do the LLM generated acoustic descriptions match the ground truth acoustic descriptions?

\textbf{RQ3:} Combining RQ1 and RQ2, do the emotion pairs that the omni-LLM most frequently misclassified also systematically exhibit discrepancies in their acoustic feature descriptions (e.g., LLM predicts ``high volume'', but ground truth is ``low volume'')?
\subsection{RQ1: Confusion Analysis}
\textbf{IEMOCAP} The most notable confusion occurred when angry was misclassified as frustrated (61\% error rate).
Other significant misclassifications included cases where the omni-LLM defaulted to neutral: frustrated predicted as neutral (23\% error rate), excited predicted as neutral (22\% error rate), sad predicted as neutral (25\% error rate). Our findings are consistent with previous works \citep{hu-etal-2023-supervised,wu2024beyond,yun-etal-2024-telme}, although we note we have a higher error rate on angry misclassified as frustrated. 

\textbf{MELD} The most significant misclassifications in MELD occurred when sadness was misclassified as neutral, with a 64\% error rate. Other notable confusions also are primarily due to the omni-LLM frequently defaulting predictions to neutral: joy predicted as neutral (42\% error rate), anger predicted as neutral (41\% error rate), surprise predicted as neutral (44\% error rate). Our findings again align with previous research \citep{hu-etal-2023-supervised,yun-etal-2024-telme}, where both noticed neutral defaulting for the MELD corpus.

\textbf{Summary} Our confusion analysis reveals consistent patterns among both IEMOCAP and MELD, namely the neutral defaulting. However, across both corpora, we closely align with confusion patterns observed in fine-tuned models. 
\subsection{RQ2: Acoustic Descriptions}
We begin by defining our setup and motivation. \citet{wu2024beyond} extracted raw audio features from the speech files, namely volume and volume variation, pitch and pitch variation, and speaking rate. From these features, the authors binned by threshold and converted these into natural language descriptions (e.g., if threshold is a certain value, the volume is \textit{low}). We compare our omni-LLM generated features in the ``Acoustic Analysis'' section of our prompt to these reference features, evaluating on F1.

\textbf{IEMOCAP}
Our IEMOCAP results show that average acoustic features are more accurately predicted than their variations, with average volume achieving the highest F1 score (0.68), followed by average pitch (0.62) and speaking rate (0.60). The model, however, struggles with capturing acoustic variations with substantially lower F1 scores for volume variation (0.33) and pitch variation (0.27). 

\textbf{MELD} Our MELD results achieve the highest F1 score for pitch(0.69), followed by average volume (0.63) and speaking rate (0.62), showing relatively strong prediction of baseline acoustic features. However, the model performs substantially worse on capturing acoustic variations, with both volume and pitch variations having an identical F1 of 0.26. 

\textbf{Summary} Our findings suggest that while omni-LLMs can reasonably predict baseline acoustic levels from audio alone, they have difficulty capturing the dynamic fluctuations.

\subsection{RQ3: Acoustic Descriptions \& Emotion Confusion} 
Table~\ref{tab:divergence} shows distinctive acoustic perception patterns, or confusion patterns the omni-LLMs default to. We focus on the three main features from RQ2: volume, pitch, and speaking rate.
We split the patterns into three categories: \textit{$\{$higher/lower/same$\}$}, indicating the predicted feature is \textit{$\{$higher/lower/same$\}$} compared to the reference feature. For example, if the acoustic reference says \textbf{low volume, low pitch, slow speaking rate}, and the LLM acoustic description says \textbf{high volume, high pitch, and fast speaking rate}, then the LLM predictions are all considered \textit{higher}. 

\textbf{IEMOCAP} When \textit{angry} is misclassified as \textit{frustrated}, the model frequently overestimates acoustic features (approximately 50\% for volume, pitch, and rate) or maintains similar values (40-44\%), suggesting a systematic perception gap. For \textit{frustrated} misclassified as \textit{neutral}, the model predominantly maintains similar pitch perception (67\%) while showing a tendency to underestimate speaking rate (53.4\%). The \textit{happy}→\textit{excited} confusion shows the clearest pattern, with strong tendencies to overestimate pitch (69.2\%) and speaking rate (71.2\%), indicating that the model perceives happy utterances with heightened acoustic characteristics.

\textbf{MELD} Our results for MELD further support the ``neutral defaulting'' pattern observed in RQ1. Across all three major confusion pairs (\textit{joy}→\textit{neutral}, \textit{sad}→\textit{neutral}, \textit{anger}→\textit{neutral}), we observe remarkably high percentages of unchanged acoustic features, particularly for speaking rate (82-92\%). This suggests that the model's tendency to default to neutral classifications also  directly corresponds to its failure to detect acoustic variations that distinguish emotional speech. These results starkly contrast IEMOCAP, where acoustic misperceptions vary by emotion pair. 

\textbf{Summary} Our analysis reveals that omni-LLMs' emotion confusion patterns are closely tied to specific acoustic feature misperceptions. For IEMOCAP, these misperceptions vary by emotion pair, while for MELD, the neutral defaulting corresponds to a failure to detect distinctive acoustic variations. Our findings emphasize a perception gap in omni-LLMs, implying that improving acoustic feature perception may further enhance zero-shot emotion recognition capabilities.

\section{Conclusion}
We present OmniVox, the first systematic zero-shot emotion recognition evaluation using four different omni-LLMs (three closed, one open source). Our analysis demonstrates that omni-LLMs not only rival, but even sometimes surpass fine-tuned audio models on the emotion corpora of IEMOCAP and MELD. We present multiple prompting strategies and find that adding acoustic analysis helps omni-LLMs. We find that for IEMOCAP, adding context helps, whereas for MELD, the results are not as consistent, particularly for GPT-4o-Audio. We conclude with an error analysis, particularly focusing on the generated acoustic descriptions. We find that omni-LLMs do not perfectly match with ground truth descriptions, and furthermore, confusion patterns are closely tied to acoustic analysis mismatches.

\bibliography{colm2025_conference, anthology}
\bibliographystyle{colm2025_conference}

\appendix
\section{Limitations \& Future Work}
OmniVox presents the first systematic use and evaluation advancement in zero-shot ERC using omni-LLMs. However, we acknowledge some limitations of our work.

\textbf{Reliance on Omni-LLMs} Our approach relies on pre-trained omni-LLMs, particularly three closed omni-LLMs (GPT-4o-audio, Gemini-2.0-Flash, Gemini-2.0-Flash-Lite). We attempt to address this by making our experiments as reproducible as possible and evaluating on one open omni-LLM (Phi-4-Multimodal-Instruct).

\textbf{Performance Metrics} The zero-shot nature of OmniVox, although in some instances outperforming fine-tuned audio-only models, falls short on other modalities (text, text+audio). We emphasize that our paper primarily focuses on the audio modality, and that omni-LLMs may not perform as well as text-only LLMs. We do intend to further explore this performance gap in future work. 

\textbf{Reliance on Two Corpora} We do note that our work relies on only two corpora: IEMOCAP and MELD. While these two corpora are widely used and benchmarked by the many previous works, they may not generalize to real-world emotion detection and classification. 

\textbf{Future Work} We hypothesize that our work may transfer to other cognitive state tasks (e.g. sentiment, belief, intent). This paper provides a framework and initial exploration into one of many cognitive state tasks (emotion). In future work, we will use the insights and discoveries from this work to apply to other cognitive state tasks. 

\section{Reproducibility statement}
\label{appendix:details}

\textbf{Code.} We will release all code in the final version of this paper. 

\textbf{APIs.} For Gemini-2.0-Flash, Gemini-2.0-Flash-Lite we used the Google Gemini API. For our GPT-4o-Audio experiments, we used the OpenAI API. To maximize reproducibility, we set our temperature value to 0.0.

\textbf{Runs.} Due to the high API cost (particularly for GPT-4o-audio) and computational overhead with running Phi-4, all results reported are an average over 2 runs. For IEMOCAP, the average standard deviation is 0.25\%, and for MELD, the average standard deviation is 0.21\%.

\textbf{Phi-4-Multimodal-Instruct. } We run Phi-4-Multimodal locally. We use the default hyperparameter values (max\_new\_tokens=1000) and temperature=0.0. We run the model locally on a shared H100 GPU. Each experiment took approximately 1 hour to run. 

\section{Examples}
In this section, we showcase four correctly classified examples from our omni-LLMs. We show two examples from IEMOCAP and two from MELD.
\label{appendix:examples}
\begin{tcolorbox}[fonttitle=\small\bfseries,
fontupper=\scriptsize\sffamily,
fontlower=\fon{put},
enhanced,
left=2pt, right=2pt, top=2pt, bottom=2pt,
title=Example 1: IEMOCAP. Gemini-2.0. With Context]
\begin{lstlisting}[language={}, breaklines=true]
CONVERSATION:
Speaker 0: All the time, every day, every day. They have e-mail over there and stuff like that, right? I can send you pictures.

Speaker 1: "Yeah, as much as I can."

Speaker 0: "Yeah."

Target Utterance: I love you

LLM OUTPUT:
Conversational Context: The speaker is reassuring someone, possibly after an apology. He mentions calling someone, figuring something out, and sending pictures via email.

Acoustic Analysis: The target audio has a low pitch and a slow pace. The tone is subdued and quiet. There's a slight trailing off at the end.

Reasoning: The low pitch, slow pace, and quiet tone suggest sadness or resignation. The trailing off at the end further reinforces this. While the context suggests reassurance, the acoustic features of the target audio point towards sadness.

Label: sad
\end{lstlisting}
\end{tcolorbox}

\begin{tcolorbox}[fonttitle=\small\bfseries,
fontupper=\scriptsize\sffamily,
fontlower=\fon{put},
enhanced,
left=2pt, right=2pt, top=2pt, bottom=2pt,
title=Example 2: IEMOCAP. GPT-4o-Audio. No Context]
\begin{lstlisting}[language={}, breaklines=true]
TARGET UTTERANCE: All right.  All right, just listen.

Acoustic Analysis: The tone of the speaker is firm and commanding. The pitch is relatively steady, without significant fluctuations. The rhythm is controlled, and the intensity is moderate, suggesting a level of assertiveness.

Reasoning: The firm and commanding tone, along with the steady pitch and controlled rhythm, indicates that the speaker is trying to assert control or make a point. The moderate intensity suggests that the speaker is not overly emotional but is focused on being heard and understood.

Label: frustrated
\end{lstlisting}
\end{tcolorbox}

\begin{tcolorbox}[fonttitle=\small\bfseries,
fontupper=\scriptsize\sffamily,
fontlower=\fon{put},
enhanced,
left=2pt, right=2pt, top=2pt, bottom=2pt,
title=Example 3: MELD. Gemini-2.0. With Context]
\begin{lstlisting}[language={}, breaklines=true]
CONVERSATION:
Speaker 0: Car's waiting downstairs, I just wanted to drop off these copies of my book for your friends. Anything you want from Lisbon?

Speaker 1: No, just knowing you're gonna be there is enough.

Speaker 0: Alright, well, be good, I love you.

Target Utterance: You kissed my best Ross! ...Or something to that effect."

LLM OUTPUT: 
Conversational Context: A woman is leaving for Lisbon and says goodbye to a man. The target audio reveals that the woman is upset because the man kissed her best friend.

Acoustic Analysis: The target audio is spoken with a moderate speed and intensity. The pitch is slightly elevated, and the tone is accusatory and sharp. There's a noticeable emphasis on the word \"kissed\" and \"Ross,\" suggesting strong emotion.

Reasoning: The elevated pitch, accusatory tone, and emphasis on key words indicate a strong negative emotion. While it could be sadness or anger, the accusatory tone and sharp delivery point more towards anger. There's no indication of fear, surprise, joy, disgust, or neutrality.

Label: anger
\end{lstlisting}
\end{tcolorbox}

\begin{tcolorbox}[fonttitle=\small\bfseries,
fontupper=\scriptsize\sffamily,
fontlower=\fon{put},
enhanced,
left=2pt, right=2pt, top=2pt, bottom=2pt,
title=Example 4: MELD. GPT-4o-Audio. No Context.]
\begin{lstlisting}[language={}, breaklines=true]
Target Utterance: Hey, Joey, could you pass the cheese?

LLM OUTPUT: 
Acoustic Analysis: The tone of the speaker is light and casual, with a steady pitch and moderate rhythm. The intensity is low, indicating a relaxed and calm demeanor.

Reasoning: The speaker's tone and pitch do not show any signs of heightened emotion such as anger, joy, or sadness. The steady rhythm and low intensity suggest a neutral state, without any emotional extremes.

Label: neutral
\end{lstlisting}
\end{tcolorbox}

\section{Omni-LLM Modalities}
Each Omni-LLM that we use has different input and output modalities. Table~\ref{tab:model-modalities} succinctly summarizes them.
\begin{table}[!htbp]
    \centering
    \begin{tabular}{l|ccc|ccc}
        \toprule
        & \multicolumn{3}{c|}{Input Modality} & \multicolumn{3}{c}{Output Modality} \\
        \cmidrule(lr){2-4} \cmidrule(lr){5-7}
        Model & Text & Audio & Image & Text & Audio & Image \\
        \midrule
        Gemini-2.0-Flash & \cmark & \cmark & \cmark & \cmark & \xmark$^\dagger$ & \cmark \\
        Gemini-2.0-Flash-Lite & \cmark & \cmark & \cmark & \cmark & \xmark & \xmark \\
        Phi-4-Multimodal-Instruct & \cmark & \cmark & \cmark & \cmark & \xmark & \xmark \\
        GPT-4o-Audio & \cmark & \cmark$^*$ & \xmark & \cmark & \cmark & \xmark \\
        \bottomrule
    \end{tabular}
    \caption{Omni-model capabilities by input and output modalities. $^\dagger$denotes that the modality is not available in the API yet. $^*$ denotes that the modality is mandatory to get an API response.}
    \label{tab:model-modalities}
\end{table}

\section{Prompts}
\label{appendix:prompts}
\begin{tcolorbox}[fonttitle=\small\bfseries,
fontupper=\scriptsize\sffamily,
fontlower=\fon{put},
enhanced,
left=2pt, right=2pt, top=2pt, bottom=2pt,
title=CoT Prompt]
\begin{lstlisting}[language={}, breaklines=true]
Please listen to this audio clip and analyze the speaker's emotional state based solely on acoustic features (tone, pitch, speed, intensity, etc.).


After listening to the audio, classify the emotion as one of:
- anger
- joy
- sadness
- surprise
- fear
- disgust
- neutral

Your emotion classification should be based on the acoustic properties of the audio.

{OPTIONAL_TEXT_INSTRUCTION}
{OPTIONAL_TEXT_CONTEXT}

Output Format:
Conversational Context: Brief summary of the interaction based on the audio clips
Acoustic Analysis: Detailed analysis of vocal cues in the audio (tone, pitch, rhythm, intensity)
Reasoning: Step-by-step justification for your emotion classification
Label: The emotion of the audio where emotion is one of anger, joy, sadness, surprise, fear, disgust, or neutral.
\end{lstlisting}
\end{tcolorbox}

\begin{tcolorbox}[fonttitle=\small\bfseries,
fontupper=\scriptsize\sffamily,
fontlower=\fon{put},
enhanced,
left=2pt, right=2pt, top=2pt, bottom=2pt,
title=Acoustic Prompt]
\begin{lstlisting}[language={}, breaklines=true]
You'll hear several audio clips from a conversation.

The first few clips provide conversational context. For the FINAL audio clip labeled 'TARGET', analyze the speaker's emotional state based on acoustic features (tone, pitch, speed, intensity).
- anger
- joy
- sadness
- surprise
- fear
- disgust
- neutral

Output Format:
Acoustic Analysis: Brief analysis of vocal cues (tone, pitch, rhythm, intensity).
Label: The emotion of the TARGET audio clip (anger, joy, sadness, surprise, fear, disgust, or neutral).
\end{lstlisting}
\end{tcolorbox}

\begin{tcolorbox}[fonttitle=\small\bfseries,
fontupper=\scriptsize\sffamily,
fontlower=\fon{put},
enhanced,
left=2pt, right=2pt, top=2pt, bottom=2pt,
title=Minimal Prompt]
\begin{lstlisting}[language={}, breaklines=true]
You'll hear several audio clips from a conversation.

The first few clips provide conversational context. For the TARGET audio clip, please classify its emotion as one of:
- anger
- joy
- sadness
- surprise
- fear
- disgust
- neutral

Label: <emotion>
\end{lstlisting}
\end{tcolorbox}

\begin{tcolorbox}[fonttitle=\small\bfseries,
fontupper=\scriptsize\sffamily,
fontlower=\fon{put},
enhanced,
left=2pt, right=2pt, top=2pt, bottom=2pt,
title=No Context Prompting Style]
\begin{lstlisting}[language={}, breaklines=true]
Please listen to this audio clip{text_context_simple} and analyze the speaker's emotional state based solely on acoustic features (tone, pitch, speed, intensity, etc.).

After listening to the audio, classify the emotion as one of:
- anger
- joy
- sadness
- surprise
- fear
- disgust
- neutral

Your emotion classification should be based on the acoustic properties of the audio.

Output Format:
Acoustic Analysis: Detailed analysis of vocal cues in the audio (tone, pitch, rhythm, intensity)
Reasoning: Step-by-step justification for your emotion classification
Label: The emotion of the audio where emotion is one of anger, joy, sadness, surprise, fear, disgust, or neutral.
Label: <emotion>
\end{lstlisting}
\end{tcolorbox}

\end{document}